\documentclass[11pt,a4paper]{article}
\usepackage[hyperref]{acl2020}
\usepackage{times}
\usepackage{latexsym}

\usepackage{CJKutf8}
\usepackage{bm}
\usepackage{multirow}
\usepackage{amssymb}
\usepackage{amsmath}
\usepackage{graphicx}

\usepackage{microtype}

\aclfinalcopy 

\title{Pretrained Language Models for Document-Level Neural Machine Translation}

\author{Liangyou Li\and Xin Jiang\and Qun Liu \\
  Huawei Noah's Ark Lab\\
  \texttt{liliangyou@huawei.com} \\}

\date{}

\begin{document}
\maketitle
\begin{abstract}
   Previous work on document-level NMT usually focuses on limited contexts because of degraded performance on larger contexts. In this paper, we investigate on using large contexts with three main contributions: (1) Different from previous work which pertrained models on large-scale sentence-level parallel corpora, we use pretrained language models, specifically BERT \citep{bert}, which are trained on monolingual documents; (2) We propose context manipulation methods to control the influence of large contexts, which lead to comparable results on systems using small and large contexts; (3) We introduce a multi-task training for regularization to avoid models overfitting our training corpora, which further improves our systems together with a deeper encoder. Experiments are conducted on the widely used IWSLT data sets with three language pairs, i.e., Chinese--English, French--English and Spanish--English. Results show that our systems are significantly better than three previously reported document-level systems.
\end{abstract}

\section{Introduction}


Recently, document-level Neural Machine Translation (NMT) is drawing more attention from researchers studying on incorporating contexts into translation models \cite{hierrnn:wang,docnmt:cho,docnmt:concat:tiedemann,docnmt:analpha:voita,docnmt:liuyang,docnmt:hieratt,docnmt:cache:suzhou,docnmt:cache:tu}. It has been shown that nmt can be improved by taking document-level context information into consideration. However, one of the common practices in previous work is to only consider very limited contexts (e.g., two or three sentences) and therefore long dependencies in documents are usually absent during modeling the translation process. Although previous work has shown that when increasing the length of contexts, system performance would be degraded, to the best of our knowledge, none of them addresses the problem this work considers.

Given the importance of long-range dependencies \citep{transformer-xl}, in this paper we investigate approaches to take large contexts (up to 512 words in our experiments) into consideration. Our model is based on the Transformer architecture \citep{transformer} and we propose methods to narrow the performance gap between systems using different lengths of contexts. In summary, we make three main contributions:
\begin{itemize}
\item We use pretrained language models (PLMs) to initialize parameters of encoders. Different from pretrained models on large-scale sentence-level parallel corpora \citep{docnmt:cache:tu,docnmt:liuyang}, PLMs are trained on monolingual documents which are easier to obtain than bilingual corpora.
\item We propose methods to manipulate the integration of context information to control the influence of large contexts. In our experiments, these methods lead to comparable results on systems using small and large contexts.
\item We introduce a multi-task training which adds an extra task on the encoder side regularizing our model and further improving our systems together with a deeper encoder. 
\end{itemize}
Experimental results on the widely used IWSLT data sets \cite{iwslt:data} show that our final systems significantly outperform systems in previous work on three language pairs, i.e., Chinese--English (Zh-En), French--English (Fr-En) and Spanish--English (Es-En). Our results also demonstrate the necessity of PLMs and usefulness of the multi-task training and the context manipulation methods.

\begin{figure}[t]
    \centering
    \includegraphics[width=0.48\textwidth]{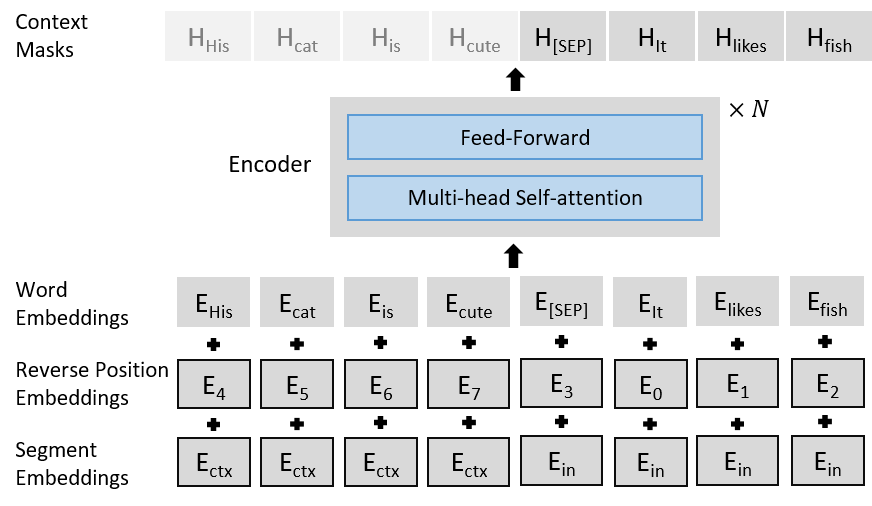}
    \caption{The proposed encoder structure. An input sentence ``{\it It likes fish}'' is concatenated with its contexts ``{\it His cat is cute}''. A separation mark ``{\it [SEP]}'' is inserted between them. Compared with the original Transformer architecture, we make the following changes: (1) {\it segment embeddings} are added to distinguish contexts from the input; (2) {\it reversed position embeddings} are introduced as an alternative of the original sequential position embeddings; (3) {\it context masks} are used during decoding to avoid attention weights on the contexts. In this figure, each ``$\mathrm{E_{*}}$'' denotes an embedding vector, while each ``$\mathrm{H_{*}}$'' denotes an output vector of the encoder.}
    \label{fig:model}
\end{figure}



\section{Document-Level NMT}\label{sec:approach}

In this paper, we consider source contexts, i.e., sentences before the current input to be translated. Following \citet{docnmt:concat:tiedemann}, the current input and its contexts are concatenated, as shown in Figure \ref{fig:model}. Instead of defining an additional hyper-parameter on the length of contexts (e.g., $k$ sentences),  we set the maximum total length of an input and its contexts to 512 words,  which is defined by the model capacity (i.e., the maximum input length).


However, incorporating such large contexts could result in unstable training and introduces much irrelevant information. To alleviate these problems, we (1) use PLMs trained on large-scale monolingual documents to initialize parameters of encoders; (2) propose a few changes of the encoder architecture to control impacts of contexts; and (3) introduce a multi-task training mechanism to regularize our model. Figure \ref{fig:model} shows the input format and encoder architecture we use in this paper.


\subsection{Pretrained Language Models}

Because large-scale parallel corpora with document boundaries are usually unavailable, researchers have tried to make use of sentence-level parallel corpora to help training a document-level NMT model \cite{docnmt:liuyang,docnmt:cache:tu}. Different from them, we use large-scale monolingual data which is much easier to obtain, e.g., from Wikipedia or other public websites. Instead of training a language model from scratch on monolingual documents, we directly use pretrained BERT models to initialize our encoder and then fine-tune our NMT model on document-level parallel corpora. BERT is chosen based on the following reasons: (1) BERT is based on the Transformer architecture considering bidirectional contexts which makes it compatible with our encoder; (2) BERT is trained over long sequences and learns relationships between them, and thus is suitable to model document-level contexts; (3) BERT codes and multilingual models are publicly available, which makes it easier to replicate our results.

\subsection{Context Manipulation}\label{sec:approach:ctxt}

Unlike previous work which uses additional components, such as context encoders or attention layers, to encode and integrate contexts, in this paper we use a single encoder on the concatenation of an input and its contexts. Although a model can be directly trained without any modification \cite{docnmt:concat:tiedemann}, we found it does not work well when large contexts are used especially without PLMs. We presume this is because large contexts introduce much more irrelevant information which would overwhelm the source sentence to be translated. To alleviate this problem, we introduce three techniques into the encoder to explicitly make a distinction between contexts and the input. 

\subsubsection{Segment Embeddings}

The concept of {\it segment embeddings} is introduced in BERT. The basic idea is that each sequence has a unique embedding so as to distinguish from other sequences. In this paper, we directly adapt the idea into the encoder and add different segment embeddings to contexts and the source sentence, respectively.


\subsubsection{Reversed Position Embeddings}


By default, position embeddings in the Transformer are assigned words by words. However, when contexts are concatenated to an input sentence, position embeddings of the source input will depend on length of the contexts which precede the source. To alleviate this, we propose to first assign position embeddings to the source input and then to its contexts, called {\it reversed position embeddings}, as illustrated in Figure \ref{fig:model}, which keeps the positional representations of source sentences stable.


To alleviate this problem, we propose to first assign position embeddings to the source input and then to its contexts, called {\it reversed position embeddings}, as illustrated in Figure \ref{fig:model}, which keeps the positional representations of source sentences stable.


\subsubsection{Context Masks}

\citet{docnmt:concat:tiedemann} has shown that directly augmenting an input with its contexts improves translation quality in RNN-based models. However, they only use the immediately previous sentence as contexts in experiments. When larger contexts are used, this kind of method does not work well because large contexts could result in unstable training and it would be more challenging for the model to learn appropriate attention weights to distinguish contexts and inputs. Therefore, in this paper we add {\it context masks} to avoid the decoder attending to the contexts as we presume representations of the source part are already context-aware through the underlying self-attention in the encoder.


\subsection{Multi-task Training}


Inspired by \citet{pretrain:seq2seq}, we introduce an extra task on the encoder side to avoid the model overfitting our training corpus. The task we use is called masked language model (MLM) prediction which is also used to train BERT. When MLM is considered, the training objective becomes: 
\begin{align}
    \hat{\theta} = \underset{\theta}{\mathrm{argmax}} \sum_{(X,C,Y)\in D} & \log P(Y|S)\nonumber\\ & + \sum_{k\in M} \log P(s_k|S),\nonumber
\end{align}
where $S$ is the extended input by combining the input $X$ and context $C$ but with some words randomly masked (about $16\%$ with a maximum number of 20 words), $M$ is the set of masked positions, $s_k$ is the real word form at position $k$.


\section{Experiments}\label{sec:exp}


\subsection{Data sets and Settings}

We conduct experiments on the widely used IWSLT data sets with three language pairs: Zh-En, Fr-En and Es-En, each of which contains around 0.2M sentence pairs. We use {\it dev2010} for development. {\it tst2010-2013} (Zh-En), {\it tst2010} (Fr-En) and {\it tst2010-2012} (Es-En) are used for testing. 


\begin{table*}
\centering
\caption{BLEU scores and increment over the best previous approach on three language pairs. ``{\it +Large Context Manipulation}'' denotes the three context manipulation methods on large contexts. ``{\it+Encoder-12}'' means to increase the number of encoder layers to 12. ``{\it +MLM}'' means adding the MLM prediction task into the training objective. The best BLEU scores are in bold.}\smallskip
\label{tab:results}
\begin{tabular}{|c|l|l|l|l|}
\hline & \bf Systems & \bf Zh--En & \bf Fr--En & \bf Es--En \\ \hline
\multirow{3}{*}{\it\small previous}& \cite{docnmt:cache:tu} & 17.32 & - & 36.46 \\
& \cite{docnmt:hieratt} & 17.79 & - & 37.24 \\
& \cite{docnmt:liuyang} & - & 36.04 & - \\ \hline
\multirow{3}{*}{\it\small our work} & Baseline & 17.31 & 35.33 & 37.01 \\
& +BERT +Large Context Manipulation & 20.10 (+2.31) & 37.46 (+1.42) & 39.53 (+2.29) \\
& \hspace{5mm}+Encoder-12L & 20.59 (+2.80) & 38.10 (+2.06) & 40.08 (+2.84) \\
& \hspace{10mm}+MLM  & \bf 20.72 (+2.93)  & \bf 38.76 (+2.72)  & \bf 40.31 (+3.07)  \\
\hline
\end{tabular}
\end{table*}



The size of our baseline NMT model follows that of BERT-base models. We train models up to 300K steps with each batch around 3072 source or target tokens. Adam \cite{adam} is used to optimize parameters with the same learning rate as the original Transformer. We directly use pretrained Chinese and multilingual BERT models\footnote{\url{https://github.com/google-research/bert}} to initialize encoders for Zh, Fr and Es, respectively. Beam search is used with a beam width of 4 and a length penalty \cite{gnmt} of 1.



\subsection{Overall Results}



In this section, we compare our document-level NMT systems with three existing approaches \citep{docnmt:cache:tu,docnmt:hieratt,docnmt:liuyang}. To ensure a fair comparison, all systems are based on the Transformer architecture, and our translations are processed and evaluated following these papers as well.

Table \ref{tab:results} shows the overall evaluation results on test sets in all three language pairs. Our systems with BERT and context manipulation methods achieve significantly better BLEU scores than previous work. Specifically, gains on Zh-En, Fr-En and Es-En are 2.80 BLEU, 2.06 BLEU and 2.84 BLEU, respectively. We also found using a deeper (12 layers) encoder improves systems compared to a shallower (6 layers) encoder (by up to 0.49 BLEU on Zh-En, 0.64 BLEU on Fr-En, and 0.55 BLEU on Es-En, respectively). When we introduce the MLM task into the deep model, the systems is further improved.


\begin{table}[t]
\centering
\caption{BLEU scores and increment over the baseline on dev2010 in ablation study. ``{\it +Large Context}'' means large contexts are simply concatenated with inputs following \citet{docnmt:concat:tiedemann}. {\it CtxMask}, {\it SegEmb} and {\it RevPos} denote Context Masks, Segment Embeddings and Reverse Position Embeddings, respectively.}\smallskip
\label{tab:ablation}
\begin{tabular}{|l|l|}
\hline \bf Systems & \bf Zh--En \\ \hline
Baseline & 12.19 \\
+BERT & 13.23 (+1.04)\\
\hspace{5mm} +Large Context  & 14.54 (+2.35) \\
\hspace{8mm} +CtxMask & 14.87 (+2.68)\\
\hspace{11mm} +SegEmb & 15.00 (+2.81)\\
\hspace{14mm} +RevPos & 15.30 (+3.11)\\
\hline
\end{tabular}
\end{table}

\subsection{Ablation Study}\label{sec:exp:ablation}

Despite the overall improvements shown in Table \ref{tab:results}, it would be interesting to know the contribution of each method we applied. Therefore, in this section, we conduct ablation study with results shown in Table \ref{tab:ablation}. We found that simply using the BERT to initialize parameters of the encoder improves the baseline system by 1.04 BLEU. When contexts are concatenated with source sentences which are then directly taken as inputs of the model without any changes in the network structure, the system (i.e., the one with ``{\it +Large Context}'') is further improved by 1.31 BLEU. Finally, when the three context manipulation methods are integrated, the system achieves the best BLEU score.

\subsection{Context Length}

\begin{table}[t]
\centering
\caption{BLEU scores and increment over the baseline on dev2010 when context length is varied. {\it Small context} denotes the immediately previous sentence. {\it +Manipulation} means the three context manipulation techniques.}\smallskip
\label{tab:ctxt} 
\begin{tabular}{|l|l|}
\hline \bf Systems & \bf Zh--En \\ \hline
Baseline & 12.19 \\
\hspace{2mm} +Small Context  & 12.29 (+0.1) \\
\hspace{2mm} +Large Context & Diverge \\
+BERT & 13.23 (+1.04) \\
\hspace{2mm} +Small Context  & 15.54 (+3.35) \\
\hspace{2mm} +Large Context & 14.54 (+2.35) \\
+BERT +Manipulation & - \\
\hspace{2mm} +Small Context & 15.50 (+3.31) \\
\hspace{2mm} +Large Context & 15.30 (+3.11) \\
\hline
\end{tabular}
\end{table}


Table \ref{tab:ctxt} shows results of varying context length. We found that NMT models with (especially large) contexts considered do not work well without pretraining. When parameters are initialized with BERT, both small and large contexts bring significant improvements even without using the three manipulation methods. This suggests the importance of pretraining when document-level parallel corpora are in small-scale and is consistent with findings in previous work \cite{docnmt:cache:tu,docnmt:liuyang}. However, a difference is that they pretrained models on sentence-level parallel corpora, which we think could help to further improve our systems.

Another finding is that using smaller contexts achieves a significantly better BLEU score (+1.0) than larger contexts, similar to \citet{docnmt:liuyang,docnmt:hieratt}. However, when our manipulation methods are applied, the system with large contexts is further improved resulting a narrowed gap (0.2 BLEU difference) between it and the system using small contexts. We also found that our manipulation methods does not improve the system with small contexts. This is expected since they are designed for controlling the influence of large contexts. Our results suggest that sophisticated manipulation on the integration of large contexts is necessary and promising to achieve a better performance.

\section{Conclusion}\label{sec:conclusion}

In this paper, we investigate document-level NMT using large contexts. We (1) use pretrained language models, i.e. BERT, to initialize the encoder; (2) propose manipulation methods to control the influence of large contexts; and (3) introduce a multi-task training mechanism for model regularization. Experiments on IWSLT data sets showed that our systems achieved the best BLEU scores compared with previous work on Chinese--English, French--English and Spanish--English. 

\bibliography{acl2020}
\bibliographystyle{acl_natbib}

\end{document}